\documentclass[sigconf, nonacm]{acmart}
\usepackage{booktabs}    
\usepackage{multirow} 
\usepackage{graphicx} 
\usepackage{epstopdf}
\usepackage{indentfirst}

\newcommand\vldbdoi{XX.XX/XXX.XX}
\newcommand\vldbpages{XXX-XXX}
\newcommand\vldbvolume{14}
\newcommand\vldbissue{1}
\newcommand\vldbyear{2020}
\newcommand\vldbauthors{\authors}
\newcommand\vldbtitle{\shorttitle} 
\newcommand\vldbavailabilityurl{URL_TO_YOUR_ARTIFACTS}
\newcommand\vldbpagestyle{plain} 

\begin{document}
\title{RATSF: Empowering Customer Service Volume Management through Retrieval-Augmented Time-Series Forecasting}

\author{Tianfeng Wang}
\affiliation{%
  \institution{Alibaba Group}
  \city{Hangzhou}
  \country{China}
}
\email{wangtianfeng.wtf@alibaba-inc.com}

\author{Gaojie Cui}
\affiliation{%
  \institution{Yuaiweiwu Tech}
  \city{Beijing}
  \country{China}
}
\email{cuigaojie@yuaiweiwu.com}

\begin{abstract}
An efficient customer service management system hinges on precise forecasting of service volume. In this scenario, where data non-stationarity is pronounced, successful forecasting heavily relies on identifying and leveraging similar historical data rather than merely summarizing periodic patterns. Existing models based on RNN or Transformer architectures may struggle with this flexible and effective utilization. To tackle this challenge, we initially developed the Time Series Knowledge Base (TSKB) with an advanced indexing system for efficient historical data retrieval. We also developed the Retrieval Augmented Cross-Attention (RACA) module, a variant of the cross-attention mechanism within Transformer's decoder layers, designed to be seamlessly integrated into the vanilla Transformer architecture to assimilate key historical data segments. The synergy between TSKB and RACA forms the backbone of our Retrieval-Augmented Time Series Forecasting (RATSF) framework. Based on the above two components, RATSF not only significantly enhances performance in the context of Fliggy hotel service volume forecasting but also adapts flexibly to various scenarios and integrates with a multitude of Transformer variants for time-series forecasting. Extensive experimentation has validated the effectiveness and generalizability of this system design across multiple diverse contexts.
\end{abstract}

\maketitle

\pagestyle{\vldbpagestyle}
\begingroup\small\noindent\raggedright\textbf{PVLDB Reference Format:}\\
\vldbauthors. \vldbtitle. PVLDB, \vldbvolume(\vldbissue): \vldbpages, \vldbyear.\\
\href{https://doi.org/\vldbdoi}{doi:\vldbdoi}
\endgroup
\begingroup
\renewcommand\thefootnote{}\footnote{\noindent
This work is licensed under the Creative Commons BY-NC-ND 4.0 International License. Visit \url{https://creativecommons.org/licenses/by-nc-nd/4.0/} to view a copy of this license. For any use beyond those covered by this license, obtain permission by emailing \href{mailto:info@vldb.org}{info@vldb.org}. Copyright is held by the owner/author(s). Publication rights licensed to the VLDB Endowment. \\
\raggedright Proceedings of the VLDB Endowment, Vol. \vldbvolume, No. \vldbissue\ %
ISSN 2150-8097. \\
\href{https://doi.org/\vldbdoi}{doi:\vldbdoi} \\
}\addtocounter{footnote}{-1}\endgroup


\section{Introduction}
\noindent
Accurate estimation of total service demand is crucial for customer service volume management in the travel industry, including hotels, airlines, attractions, and transaction-brokering apps like Fliggy, significantly affecting system costs.Underestimating by 100 service requests incurs urgent mobilization costs, equivalent to three times the labor cost for a single request; overestimation, meanwhile, leads to wasted labor costs. The travel industry presents unique challenges in service volume forecasting due to its interplay with variables like order statuses, weather, international events, and destination country policies. At Fliggy, precise forecasting is essential for planning the recruitment and training of customer service staff, as well as scheduling. These factors result in non-stationary data patterns that can introduce significant biases with traditional uni-variant time-series forecasting methods \cite{hyndman2018Holt-Winter,ARIMA,taylor2018Prophet}, which rely on periodic summaries and trend analyses.

In fact, across domains like stock market analysis and station traffic forecasting where business cycles and fluctuations play a significant role, there is a shared desire for a universal and flexible time series forecasting system design that can adeptly utilize historical sequence information to tackle intricate prediction challenges. Meanwhile, in the realm of time series forecasting(TSF), where models, whether based on Recurrent Neural Networks (RNNs) or Transformer, commonly face difficulties efficiently processing and extracting insights from vast amounts of historical data.

One naive way to solve this is to try elongating the sequence that the transformer processes. Some approach involves sampling historical data to fit within a limited context window. Specifically, the Informer \cite{zhou2021informer} algorithm samples K points from the sequence and derives a shorter Q sequence based on these sampled points. However, this approach assumes that all historical information is equally important, which may not be suitable for many time series scenarios where different data points can carry varying significance.
Perceiver \cite{jaegle2021perceiver} and similar methods opt for a different approach by mapping Query sequences to fixed lengths, reducing computation and allowing for more historical data storage. Nonetheless, they still face challenges in efficiently extracting and interpreting critical information from extended time series.

In the field of natural language processing, strides have been made to expand Transformer models' context handling. On one front, techniques like flash-attention \cite{dao2022flashattention} enhance efficiency and reduce complexity, enabling longer context processing. However, it's crucial to note that directly extending receptive fields may cause larger models to overlook significant details in lengthy inputs, like \cite{liu-etal:2023:Lost-in-the-Middle} mentioned.

On another front, Retrieval-Augmented Generation (RAG)-like methods have drawn attention for enhancing model performance by incorporating external information. NVIDIA's \cite{anonymous2024retrievalmeetsLongContext} research indicates that even when models already handle large context windows in text tasks, they can still achieve substantial performance gains by retrieving and using relevant data from external sources.

To address the above challenges, we have identified a potential approach employing the concept of retrieval augmentation (RA). Concretely, we focus on implementing two central enhancements: a knowledge base schema that can efficiently index all historical series, and a cross-attention module embedded in a transformer model to integrate historical information for pinpointing and exploiting the most predictive segments, thereby refining prediction accuracy. The synergy between TSKB and RACA constitutes the RATSF framework, which significantly improves the performance of most univariate time series forecasting tasks. Currently, RATSF has been integrated into four key service areas within Fliggy, encompassing hotel bookings and after-sales services, train ticketing, and flight reservations and modifications. This integration has led to a notable decrease in forecasting errors, from an approximate range of 15\% to approximately 8\% across all the aforementioned sectors, which has significantly reduced personnel management costs, although the exact extent of cost savings cannot be divulged for commercial reasons. 

In concise terms, our main contributions are: 

1. We present a straightforward and manageable design for a time-series knowledge base (TSKB) based on the characteristic that time series data is easy to be stored in a structured way, which significantly facilitates efficient management of historical data.

2. We introduce a versatile cross-attention module, retrieval augmented cross-attention (RACA) module, designed to integrate retrieved historical data into the forecasting process. This module is easily adaptable and can be seamlessly integrated with various time-series transformers. 

3. We raise a Retrieval-Augmented Time Series Forecasting (RATSF) framework, which decreases 7\% forecasting errors in the real service areas within Fliggy and therefore reduced huge personnel practical management costs. Extensive experiments on more publicly available datasets demonstrate that our method achieves the best performance for the univariate time series forecasting task and is general for a broader range of industrial applications.


\section{Review}

\noindent
Despite intense competition \cite{salinas2020deepar,das2023TIDE,ye2022kddgnn,han2022kddgnn}, Transformer models \cite{vaswani2017att-is-all} and their improved variants \cite{qing2021aliformer,woo2022etsformer,liu2021pyraformer,Kitaev2020Reformer:,lim2021tft} have become the mainstream choice in time-series forecasting tasks. However, the computational complexity of the original Transformer model scales as O(n²) with respect to sequence length, significantly limiting the maximum sequence length it can handle and thereby constraining its ability to use historical data. 

\textbf{Improvements in Time-Series Forecasting with Transformers.}
Numbers of works focus on enhancing the Transformer model's capability to extract temporal features, thereby improving prediction accuracy, exemplified by Fedformer's \cite{zhou2022fedformer} introduction of a frequency-augmented Attention mechanism that directly incorporates Fourier operators, a similar path taken by TimesNet \cite{wu2023timesnet}. In contrast, Autoformer \cite{wu2021autoformer} proposes an operator that decomposes time series information into trend and periodic components. These two types of solutions perform well for relatively stationary sequences but may see performance drop when dealing with highly non-stationary time series data.

Another core strategy involves learning a robust representation of the time series first, which is then used to enhance the forecasting accuracy. TNC \cite{tonekaboni2021tnc} harnesses the core concept of contrastive learning and employs samples within a specific temporal neighborhood within a window as positive pairs, while treating samples from differing temporal neighborhoods as negative pairs.
Despite these improvements having collectively boosted the ability of Transformers in time-series forecasting tasks, none has fundamentally increased the Transformer's receptive field.

\textbf{Retrieve Augmented Method in NLP.}
In last two years, the Retrieval-Augmented Generation (RAG) approach \cite{guu2020rag-lm,borgeaud2022rag_llm,pasupat-2021-rag_Sematic,zhang2022rag_info_extract,shuster-2021-rag-Reduces-Hallucination,Khandelwal2020knnlm,Zuo2019rag-conversational-search} has gained widespread adoption in the field of NLP. REALM harnesses a knowledge retriever to distill information from vast corpora and thereby enhance the performance of pre-trained language models. Transformer-XL+kNN \cite{Bonetta2021Transformer-XL} incorporates a K-Nearest Neighbors algorithm to search through training data, refining dialogue generation capabilities. In the context of named entity recognition tasks, U-RaNER \cite{tan-etal-2023-damo-raner} utilizes multi-modal heterogeneous retrieval techniques to boost knowledge retrive and, by integrating retrieved knowledge into the model, strengthens its understanding of queries and improves entity recognition accuracy.

\textbf{Retrieval Augmented Method in Time-series Forecasting.}
MQ-ReTCNN \cite{yang2022mqretnn} is designed for complex time series prediction tasks involving multiple entities and variables. It employs a scoring function to compile relevant contexts from offline data, selects scored segments, and appends them to the prediction sequence.Its retrieval mechanism emphasizes leveraging historical sequences of one entity to enhance predictions about another, without directly utilizing historical data for direct prediction assistance.

ReTime \cite{Jing2022retime} creates a relation graph based on temporal closeness between sequences and employs relational retrieval instead of content-based retrieval.It does not optimally use historically similar sequences as reference points due to its inherent design limitations. Both MQ-ReTCNN and ReTime incorporate retrieval enhancement strategies but have yet to introduce a general and efficient retrieval technique specifically for single-variable time series prediction scenarios.

\begin{figure*}
      \centering
      \includegraphics[width=\linewidth]{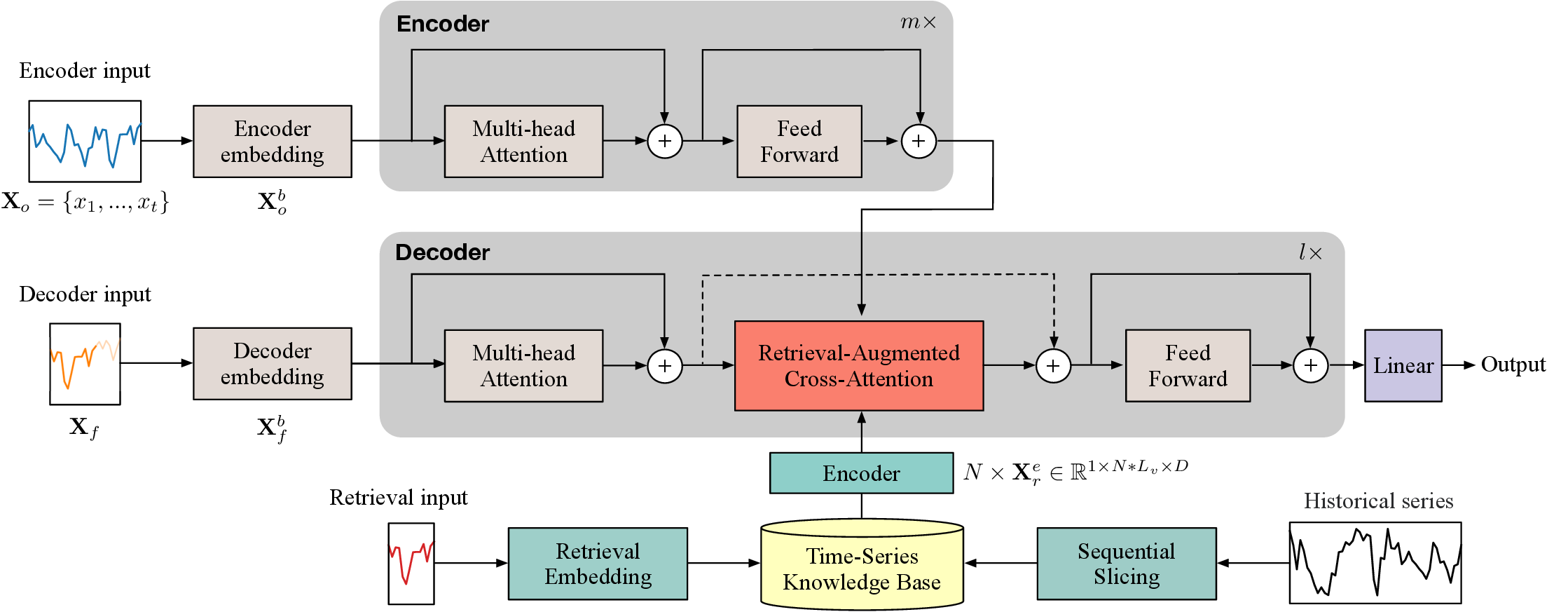}
      \caption{RATSF uses a TSKB to store and index historical sequences, alongside a transformer-based forecasting model. As illustrated in the bottom-right corner, TSKB segments the full history for efficient retrieval. To do forecasting, the Encoder fetches $\mathbf{X}_o$ from t recent time steps, and forms a retrieval sequence with d latest steps, retrieves N related sequences $\mathbf{X}_r$ from the TSKB. RACA in the Decoder then merges $\mathbf{X}_r$ and $\mathbf{X}_o$ info to deliver result. }
      \label{fig1}
\end{figure*}

\section{Method}
\noindent
\subsection{Setting \& Notations}
\label{section:setting}
Before elucidating our approach, we first clarify the setting of our problem and define some symbols that will be used throughout the text.

In actual business operations, service volume is influenced by a multitude of factors, some of which are challenging to fully represent through variables. To simplify the reasoning process and enhance the versatility of our system, we have configured our hotel service volume prediction using a uni-variant time-series forecast setup. This means that the inputs to the model consist solely of the the values of the time series and it's temporal features . Due to the requirements of our actual task—where personnel management necessitates advance preparation for recruitment, short-term scheduling, and training. Our forecasting task is designed to start from a specific time point $t$ and predict a sequence of $L_f$ data points in one go.

\textbf{Time Marking.} We take moment $t$ as the reference point, and the collection of the future time points to be forecasted is represented as ${\left[t+1, t+2, \ldots, t+l_f\right]}$, where $l_f$ denotes the length of the forecast period. Concurrently, we define the retrieve segment length as $l_r$, with the indexing sequence $\mathbf{K}$ in the TSKB having a length of $L_r$, and the length of the V sequence as $l_v$.

\textbf{Sequence Source Marking.} The subscript o denotes the original sequence, that is, the sequence $\mathbf{X_o}={x_1, ..., x_t}$ prior to moment $t$; $\mathbf{X_f}$ represents the sequence to be forecasted; $\mathbf{X_r}$ signifies the $\mathbf{V}$ sequence retrieved from the TSKB.

\textbf{Transformer Processing Marking.} The superscript is used to indicate the results after processing by various parts of the Transformer. $\mathbf{X^b_o}$ denotes the original sequence after Embedding, and $\mathbf{X^e_o}$ denotes the sequence after the Encoder. Within the Decoder, H represents the internal hidden state of the Decoder, $l$ represents the layer number, and $H^l$ indicates the hidden state processed by the $l$-th layer.

\subsection{Overview of RATSF}
The RATSF system is composed of two core components: a Time Series Knowledge Base (TSKB, detailed in Section \ref{section:tskb}) and a Transformer-based time series forecasting model. The latter has been enhanced by replacing its original decoder with our novel Retrieval Augmented Cross-Attention (RACA, described in Section \ref{section:raca}) module. The TSKB efficiently segments and archives historical time series data, establishing a precise indexing system. The transformer model, with the integration of RACA, effectively utilizes retrieved historical data to significantly enhance the accuracy of its predictions. The data flow diagram of RATSF is illustrated in Figure \ref{fig1}, providing a visual representation of the process.

\subsection{TSKB}
\label{section:tskb}
\noindent
 Unlike traditional methods that solely store and retrieve the entire original sequence, TSKB preserves the original sequence as the core content (\textbf{V}) while selects certain segments from it to construct a discriminative indexing sequence (\textbf{K}). This approach can be likened to processing a written piece where conventional methods involve direct full-text searches for desired information, which may be inefficient and less precise; whereas with TSKB, it's akin to extracting key headlines from the body of the text to serve as indices that facilitate rapid access to core information. This dual-sequence structure allows the system, when handling large-scale data, to efficiently locate and access targeted portions of the original sequence through customized index sequences, thereby significantly enhancing overall performance.

\subsubsection{Sequential Slicing}
\begin{figure}[h]
      \centering
      \includegraphics[width=\linewidth]{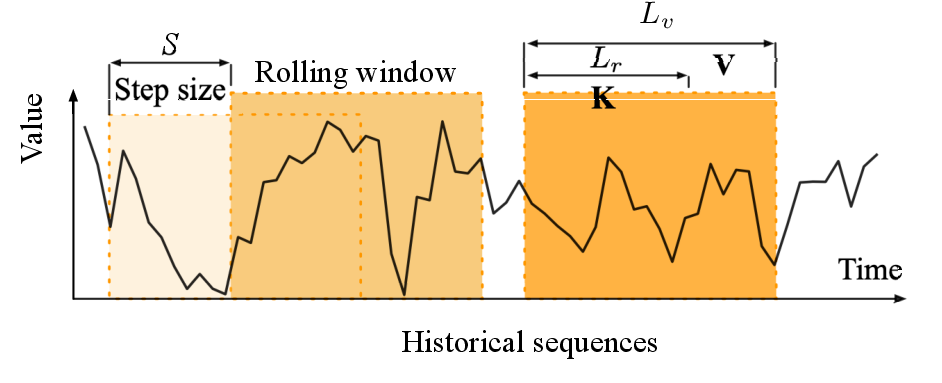}
      \caption{TSKB utilizes a rolling window of length $L_v$ to collect \textbf{V}, with an indexing segment of length $L_r$ taken from its leading part as \textbf{K}
      , and advances the window in steps of size S.}
      \label{fig_tskb}
\end{figure}

\noindent
As illustrated in Figure \ref{fig_tskb}, the content sequence \textbf{V} is obtained using a rolling window approach with a step size of $S$ and a window length of $L_v$. This approach allows us to sample all historical sequences and incorporate them into TSKB. Since prediction models only use present data to infer the future, we align indexing sequence \textbf{K} with this constraint. We extract the initial segment of each \textbf{V}, having a length of $L_r$, to serve as its index \textbf{K}. The selection of $L_v$, $L_r$ and $S$ is introduced in section \ref{section:deployment}.

\begin{figure}[h]
      \centering
      \includegraphics[width=\linewidth]{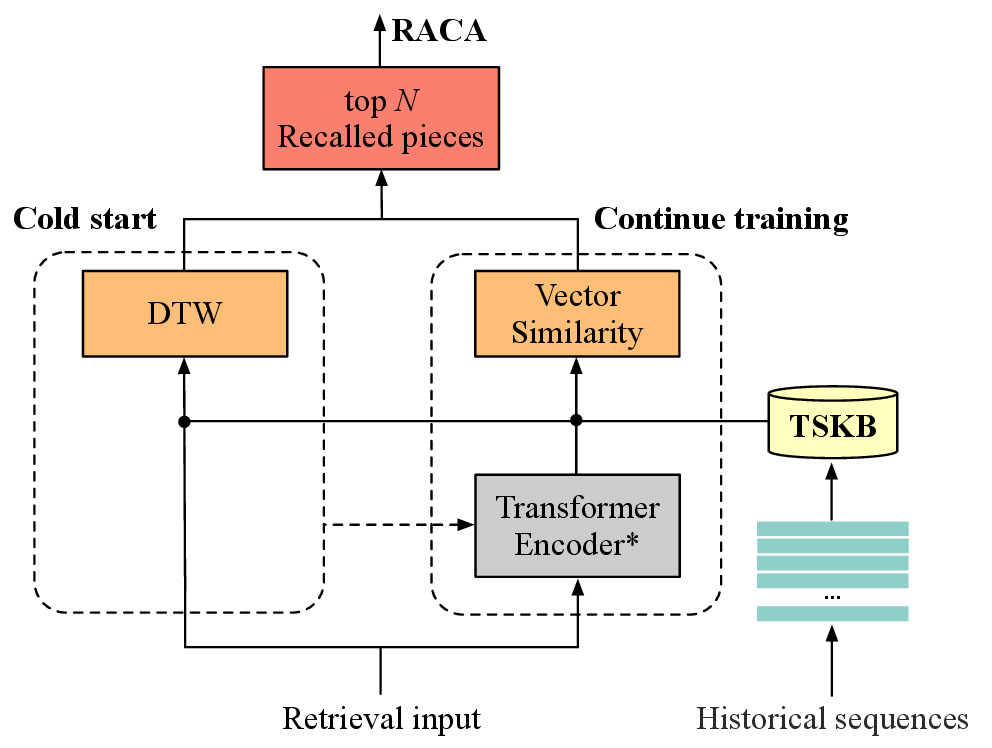}
      \caption{Within the cold start stage, we use raw sequence of retrieval input to find Top-K relevant sequences using DTW. After one epoch, we switch to Euclidean Distance for retrieval embedding to gather Top-K matches. }
      \label{fig_re_emb}
\end{figure}

\subsubsection{Embedding Learning}
\label{section:embedding learning}
\noindent
A key approach to enhancing the recall and accuracy of knowledge base retrieval is to employ a well-trained embedding vector to improve the precision of the index. In the following sections, we will focus on how to train and utilize these index embedding vectors.
Judging the quality of a retrieval embedding mainly depends on how well it captures similarities at key information points relevant to forecasting tasks. Time series data is complex due to numerous information points and the challenge of presetting comparative weights. Therefore, we use this principle: if a representation closely mirrors the important details needed for prediction, its overall performance will be better.

To ensure that the embeddings capture the essential information for forecasting, we employ the encoder from the RATSF's forecasting model, which is designed to select precise information during training, thereby enhancing forecast accuracy. Specifically, each indexing sequence \textbf{K} rom the knowledge base is processed through the encoder layer of the RATSF forecasting model to generate a retrieval embedding.

Meanwhile, in the early stages of training the forecasting model, its encoder is not yet capable of producing representations that accurately match target sequences. The model's effectiveness and learning pace are significantly correlated with its ability to retrieve historical sequences beneficial to forecasting tasks.

To avoid time and data-consuming iterations stemming from a random initialization state, we introduce Dynamic Time Warping (DTW\cite{dtw}) as an auxiliary tool during the initial phase of model training, as shown in left part of Figure \ref{fig_re_emb}. DTW is initially used for similarity-based sequence retrieval, aiding in the iterative training process of the RATSF forecasting model. After completing one epoch of training, we transition to using embeddings generated by the forecasting model itself for sequence retrieval, continuing the training until the model converges.

\begin{figure}
      \centering
      \includegraphics[width=\linewidth]{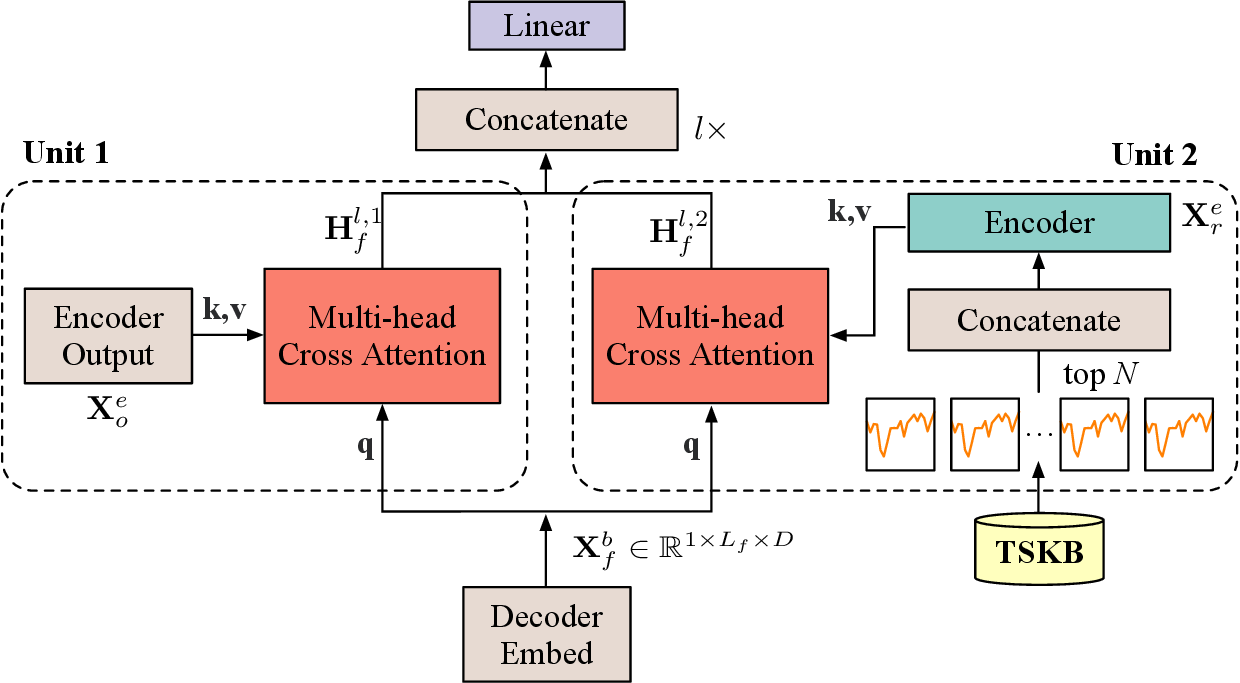}
      \caption{Each RACA has two cross-attention modules: one utilizes the Encoder output as \textbf{K},\textbf{V}, while the other employs embedded retrieved sequences as \textbf{K},\textbf{V}. Both outputs are concatenated and passed through a Linear module to reshape back to the input dimensions.}
      \label{fig_raca}
\end{figure}

\subsection{RACA}
\label{section:raca}
\noindent
As previously mentioned, Retrieval-Augmented Cross-Attention (RACA) is a module designed to be integrated into a time-series forecasting model, coupled with the TSKB. In our demonstration, we employ the vanilla Transformer as the time-series forecasting model; however, it's important to note that the RACA module is designed to be compatible and can be seamlessly plug into any transformer-based time-series model. We will specifically demonstrate this in Section \ref{section:experiments} .

The decoder consists of $l$ stacked RACA modules for inputs of length $L_f$. Each module has two parallel units: Unit 1, like in Function (1), uses $\mathbf{X}^b_f$ as Query with Key and Value from $\mathbf{X}^{e}_{o}$, outputting $\mathbf{H}^{l,1}_f$ of the same shape. Unit 2, retrieval-augment part, like in Function (2) ,also queries $\mathbf{X}^b_f$ but fetches Keys and Values from the concatenated sequence of $\mathbf{X}^{e}_{r}$, generating $\mathbf{H}^{l,2}_f$. $s$ in the Functions is the scaling factor.

Like shown in Figure \ref{fig_raca}, retrieved sequences are transformed through the encoder's embedding module, generating Ns $\mathbf{X}^{e}_{r}$, and we concatenate these N vectors into one with shape$[1, N*L_v, D]$ for later use.

\begin{gather}
\mathbf{H}^{l,1}_f = \text{softmax} \left(\frac{\mathbf{X}^b_f \mathbf{X}^{e}_{o}}{s}\right) \mathbf{X}^{e}_{o},\label{eq1} \\
\mathbf{H}^{l,2}_f = \text{softmax} \left(\frac{\mathbf{X}^b_f \mathbf{X}^{e}_{r}}{s}\right) \mathbf{X}^{e}_{r},\label{eq2} \\
\mathbf{H}^{l+1}_f = \delta \left( \text{concat} \left( \mathbf{H}^{l,1}_f,\mathbf{H}^{l,2}_f \right) \right). \label{eq3}
\end{gather}

As shown in Function \ref{eq3}, after concatenating both intermediate vectors,$\mathbf{H}^{l,1}_f$ and   $\mathbf{H}^{l,2}_f$, along the sequence dimension, forms a $[1, 2*L_f, D]$ vector, it undergoes linear transformation before being passed to the next layer.

\subsection{Deployments Procedure}
\label{section:deployment}
In this section, we briefly introduce the procedure to deploy RATSF in a new field.

\subsubsection{TSKB Initialization}
Set the initial retrieval length ($L_r$), which can be equivalent to the prediction length ($L_f$).
Set the sequence length for the V sequence ($L_v$) to be at least the sum of $L_r$ and $L_f$. This ensures adequate info for retrieval and forecast. It is suggested to set several $L_v$ values, such as $L_r$ + $L_f$, 2*$L_f$ + $L_r$, 3*$L_f$ + $L_r$, etc.,to build several TSKBs for optimal value selection later.
Set the rolling window step (stride) to 1 and collect $V$ sequences for each TSKBs. Then, Select the initial $l_r$ elements of $V$ sequences to form $K$ sequences.
\subsubsection{Identifying Optimal $L_v$}
Train diverse RATSF models on their respective TSKBs and evaluate each by prediction accuracy.Select the model with the highest prediction accuracy as the optimal RATSF model ($M^*$), and identify the corresponding V sequence length ($L_v$) as the final chosen value ($L_v^*$).

\subsubsection{Adjusting Retrieval Length $L_r$}
Adjust the retrieval length ($L_r$) to observe the impact on the performance of the optimal model (M*). The adjustment range could be from $0.5L_f$ to $L_v$, with each adjustment increment being $0.5L_f$. Record the $L_r$ value that enables M to achieve the highest prediction accuracy, denoted as $L_r^*$.

\subsubsection{TSKB and Model Optimization}
With the confirmed optimal retrieval length ($L_r*$) and V sequence length ($L_v^*$), reconfigure the TSKB. Reinitialize and train the RATSF model with the updated TSKB till converge.

\section{Experiments} 
\label{section:experiments}
\noindent
To demonstrate the superiority of our method in service volume forecasting, we first conducted experiments using the Fliggy Hotel Service Volume Dataset (FHSV), showcasing the performance enhancement of typical time-series models when integrated with RATSF. Additionally, to validate the general applicability of our approach, we extended our experiments to three other datasets. Furthermore, through a series of ablation studies, we individually assessed the impact of each key design choice in RATSF to confirm the correctness of our detailed selections.

\subsection{Experiment Setup}
\textbf{Datasets}. Given that our primary focus lies in the forecasting task within the context of customer service volume management, we initially employed the Fliggy Hotel Service Volume Dataset to substantiate the efficacy of RATSF, the detail description of the dataset is in Appendix A. Additionally, to demonstrate the performance of our approach in other contexts, we employed three classic time-series forecasting datasets: ETT\cite{zhou2021informer}, Exchange\cite{exchange-dataset} , Traffic\cite{traffic-dataset} .

\textbf{Models}.In Section \ref{section:raca},  we established that the Retrieval Augmented Cross-Attention (RACA) module is compatible with various Transformer variants. To underscore this versatility, we selected three prominent time-series domain models: Fedformer\cite{zhou2022fedformer}, Autoformer\cite{wu2021autoformer}, and NS-Transformer\cite{liu2022nstransformer}, and compared them to the vanilla Transformer. Leveraging the specialized operators within Fedformer and Autoformer that cater to time-series periodicity and trends, such as Autoformer's seasonal and trend-cyclical initializations, we have seamlessly integrated these models with the RACA module. This integration allows for the effective incorporation of their distinct sequential decomposition capabilities, enhancing the overall processing power. Furthermore, to demonstrate the advantage of our recall mechanism in the RATSF framework, we compared it with ReTime, a RAG-based model that also uses specialized retrieval techniques. Since ReTime has not released their code and their application domain is distinct from our focus on univariate time-series forecasting, we reconstructed the Relational Retrieval method based on their published paper. (It is worth noting that, as of the time of writing this paper, neither MQ-ReTCNN nor ReTime have released their code to the public.)

As shown in Figure \ref{fig1}, the training outcomes of the encoder's output significantly influence both the retrieval quality and the decoder's performance. In the main experiments, we will compare the retrieval schemes of RATSF and ReTime. Furthermore, in several subsequent experiments, we will elucidate the differences between RATSF's retrieval approach and DTW. Consequently, for the majority of the experiments in this paper, to facilitate these comparisons, we have replaced the input to RACA from $\mathbf{X}^e_r$ to $\mathbf{X}^b_r$, and we will provide an ablation study to explore the specific effects arising from this substitution.

\textbf{Forecasting Setting}. Like we claim in Section 
\ref{section:setting}, we adopt the uni-variant time-series forecasting setting. In the Fliggy's practical operations, we are required to finalize our forecasting for the upcoming week's day-by-day staffing needs one week in advance, which is why the actual prediction window period is set at 7 days, represented as 7 tokens. In the travel industry,  which Fliggy App's serving, a quarter generally defines a relatively complete business trend. Therefore, we choose an encoder window length of 98 days—exceeding 90 days and divisible by 14—which consequently results in $\mathbf{X_o}$ is 98 units long. And we concate $\left\{x_{t-14},...,x_t\right\}$ with a placeholder of length 7 to form $\mathbf{X_f}$. To match the length of $\mathbf{X_f}$, we set $L_v$=21 and $L_r$=14.
Regarding normalization, we employ unified $\mu$ and $\sigma$ values for whitening operations during the preprocessing stage and inverse normalization is carried out after forecasting.
We have applied the same data processing scheme to both the ETT, Exchange and Traffic datasets.

\textbf{Training Setting}.We have selected the Adam optimizer, with the batch size$=$64 and  max training epochs$=$10. The initial learning rate (lr) is set to 0.0001, employing a linear decay strategy where the decay parameter for the first 5 epochs is 0.9, for the last 5 epochs is 0.5. We also incorporate $L1$ regularization with $\lambda$=0.0001. Concurrently, an early stopping mechanism is activated when the loss ceases to decrease.

\textbf{Metric}.To ensure that our evaluation metrics directly reflect business performance, we first inverse normalize the forecast result as above mentioned and then compute the MSE (Mean Squared Error) and MAE (Mean Absolute Error) against the Ground Truth. The lower these two indicators are, the more accurate the prediction results prove to be. Moreover, each decrement of 1 unit in MAE signifies a corresponding reduction of 1 unit in service staff management expenditure. This configuration has been applied uniformly across all experiments.

\begin{table*}
    \centering
    \caption{Performance comparison of the Transformer models with their retrieval-based variants on four experimental datasets. 'baseline' denotes the original model without retrieval, compared to Relational Retrieva from ReTime and our approach RATSF.All results are evaluated using  MSE (Mean Squared Error) and MAE (Mean Absolute Error), and the best result is bolded.}
    \resizebox{\linewidth}{!}{
        \begin{tabular}{c|c|cccccccc}
             \toprule         
             
             \multicolumn{2}{c|}{models}&  \multicolumn{2}{c}{transformer}&  \multicolumn{2}{c}{nstransformer}&  \multicolumn{2}{c}{autoformer}&\multicolumn{2}{c}{fedformer}\\ 
             \midrule 
             
             \multicolumn{2}{c|}{metric}&mse&mae&mse&mae& mse&mae& mse&mae\\
             
             \midrule 
             \multirow{3}{*}{FHSV}& baseline&4344450.512&1349.146&4414326.000&1242.128&6096673.500&1752.798&5517485.000&1638.275\\
             & Relational Retrieval&{3737283.320}&{1276.836} & {4037294.000} & {1168.127} & {5821383.300} & {1703.328} & {5411003.700} & {1610.320}\\
             & RATSF&\pmb{3421546.027}&\pmb{1101.172} & \pmb{3823635.000} & \pmb{1070.933} & \pmb{5597297.500} & \pmb{1657.561} & \pmb{5386818.500} & \pmb{1595.657}\\
            
            \midrule 

             \multirow{3}{*}{ETTh1} & baseline & 3.187 & 1.391 & 1.501 & 0.920 & 8.651 & 2.264 & 4.159 & 1.597 \\
             & Relational Retrieval&{2.876}&{1.206} & {1.465} & {0.900} & {8.203} & {2.197} & {3.012} & {1.572}\\
             & RATSF&\pmb{2.366} &\pmb{1.180} &\pmb{1.407} &\pmb{0.886} &\pmb{7.300} &\pmb{2.113} &\pmb{3.952} &\pmb{1.565}  \\

            \midrule 

             \multirow{3}{*}{Exchange} & baseline & 0.00165 & 0.03134 & 0.00014 & 0.00922 & \pmb{0.00058} & \pmb{0.01906} & 0.00172 & 0.03493 \\
             & Relational Retrieval&{0.00143}&{0.02845} & \pmb{0.00013} & {0.00920} & {0.00069} & {0.02013} & {0.00102}  & {0.02821} \\
             & RATSF & \pmb{0.00068} &\pmb{0.02012} &\pmb{0.00013} &\pmb{0.00912} & 0.00067 &  0.02000 &\pmb{0.00068} &\pmb{0.02084} \\

             \midrule 

            \multirow{3}{*}{Traffic} & baseline & 0.00007 & 0.00564 & \pmb{0.00006} & 0.00569 & 0.00008 & 0.00643 & 0.00012 & 0.00876 \\
            & Relational Retrieval&{0.00007}&{0.00534} & \pmb{0.00006} & {0.00512} & \pmb{0.00007} & {0.00601} & {0.00009} & {0.00718}\\
             & RATSF & \pmb{0.00006} &\pmb{0.00498} &\pmb{0.00006} &\pmb{0.00472} & \pmb{0.00007} &  \pmb{0.00514} &\pmb{0.00007} &\pmb{0.00650} \\
             
             \bottomrule 
    \end{tabular}
    }
    \label{table1}
\end{table*}

\subsection{Main Restult}
\noindent
Table \ref{table1} displays the prediction accuracy of the Transformer and its variants on four datasets, comparing performance without RA, with ReTime's RA, and with our RATSF approach. In the Fliggy Hotel Customer Service Volume Dataset forecasting scenario, the Transformer model's MAE loss was reduced by 5.34\% with ReTime's Relational Retrieval and by 18\% after incorporating RATSF. This improvement could potentially translate to a reduction of approximately 200 in redundant personnel costs in practical management operations. A horizontal analysis of the first row in Table 1 reveals that all compared Transformer-based time-series model variants experienced a significant reduction in MAE after employing Retrieval-Augmented (RA) strategies. Specifically, the adoption of RATSF resulted in respective decreases of 14\%, 5\%, and 4\%. In contrast, the application of ReTime's retreval method led to improvements of 5.9\%, 2.8\%, and 1.7\% compared to their original designs without RA. These results clearly demonstrate two conclusions: first, effective RA strategies can enhance the performance of time-series Transformer models; second, RATSF exhibits a significant advantage in its retrieval strategy.

Furthermore, cross-comparison reveals that Autoformer and FedFormer yield slightly inferior results compared to the Transformer, mainly due to the stronger non-stationarity inherent in customer service volumes data within the hotel industry. As an illustrative example, while Mondays typically exhibit similar cyclical characteristics, actual service volume during Mondays preceding a short holiday can surge significantly compared to regular ones. In such instances, historical data from periods close to other short holidays provide more valuable insights than simple temporal periodicity and short-term trends; specific case studies will be showcased later.

Additionally, across the ETT, Exchange and the Traffic dataset, the adoption of RATSF design in the Transformer and its time-series optimized variants led to noticeable reductions in MAE metrics. Moreover, among these, the NSTransformer consistently outperforms the other models, likely because its mean-adapted feature proves effective in a broad range of time-series prediction domains.

\subsection{Ablation Study}
\noindent
We confirm the effectiveness of several design choices in this section.
Section \ref{section:raca_design} emphasizes the superiority of RACA's approach in integrating historical data for improved forecasting. Section \ref{section:exp_retrieve_emb} demonstrates the advantages of utilizing model encoder for recalling historical pieces. Section \ref{section:dtw} demonstrates the benefits of incorporating DTW into the training process.

\begin{figure*}[h]
      \centering
      \includegraphics[scale=1.5]{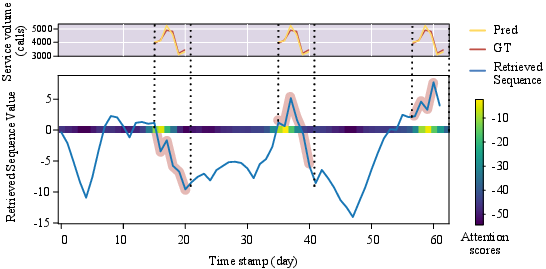}
      \caption{This figure shows RACA's use of retrieved sequences for a random forecast sequence $\mathbf{X_f^*}$.The lower half features the top 3 similar sequences of $\mathbf{X_f^*}$, concatenated along the time axis to form RACA's input. Attention weights at the first forecast point ($x_{t+1}$) are marked by data bars, with yellow and green indicating RACA's focus on downward-curving segments. The upper half replicates prediction results (yellow "pred" lines for $x_{t+1}...x_{t+f}$) and true values (orange "gt" lines), aligned with the corresponding final $L_f$ lengths of the retrieved sequences. This visual comparison highlights the model's pattern recognition, with RACA's focus areas correlating to the predictions' inflection at $x_{t+1}$.For this sample, the model notably focuses on segments indicating ascent and predicts a subsequent increase in value.}
      \label{fig_vis_ca}
\end{figure*}

\begin{table}[h] 
\centering
\caption{Ablation of the specifically designed structures within RACA on FHSV scenario.}
\begin{tabular}{cccc}
    
    \toprule         

    dataset & methods & \multicolumn{2}{c}{metric} \\
    
     \midrule 
     
     &  & mse & 4344450.512 \\
     & \multirow{-2}{*}{baseline} & mae & 1349.146 \\
     \cline{2-4} 
     
     & baseline + & mse & 4071384.750 \\
     & Design One & mae & 1226.850 \\
     \cline{2-4} 
     
     &  & mse & \bf 3421546.027 \\
    \multirow{-6}{*}{FHSV} & \multirow{-2}{*}{baseline + RACA} & mae & \bf 1101.172 \\
     
    \bottomrule 
\end{tabular}
\label{table2}
\end{table}

\subsubsection{Integrate Historical Sequences with RACA}
\label{section:raca_design}
\noindent
Based on the Transformer architecture, two designs are considered for integrating historical pieces into the forecasting model: Design One represents a straightforward approach where historical pieces are combined with the forecasting context inputs within the encoder; Design Two employs our advanced RACA design that integrate historical pieces using cross-attention mechanisms within the decoder. 

With the Fliggy Hotel Service Volume Dataset, we contrasted the performance of three methods: No historical pieces used(baseline experiment), Strategy One, and Strategy Two (RACA). The outcomes in Table \ref{table2} demonstrate that the RACA design excels, with its MAE notably lower than both the baseline and Strategy One.

A deeper examination disclosed that the Cross Attention mechanism in the RACA design efficiently guides the forecast sequence to concentrate on past trends analogous to the current scenario. In Figure \ref{fig_vis_ca}, we illustrate how retrieved sequences influence the predicted values for the first forecast point of a specific sequence through RACA's cross-attention mechanism. The lower half features the top 3 similar sequences, concatenated along the time axis to form RACA's input. Attention weights at the first forecast point ($x_{t+1}$) are marked by data bars, with yellow and green indicating RACA's focus on downward-curving segments. The upper half replicates prediction results (yellow "pred" lines for $x_{t+1}...x_{t+f}$) and true values (orange "gt" lines), aligned with the corresponding final $L_f$ lengths of the retrieved sequences. This visual comparison highlights the model's pattern recognition, with RACA's focus areas correlating to the predictions' inflection at $x_{t+1}$.

\begin{table}[h] 
\centering
\caption{Comparison of several variants of retrieval representation schemes on the FHSH dataset.}
\resizebox{\linewidth}{!}{
    \begin{tabular}{cccc}
    \toprule         

    stage & methods & \multicolumn{2}{c}{metric} \\
    
    \midrule 
    
     &  & mse & 4344450.512 \\
     & \multirow{-2}{*}{baseline} & mae & 1349.146 \\
      \cline{2-4} 

     &  & mse & 3787191.250 \\
     & \multirow{-2}{*}{baseline + DTW} & mae & 1226.690 \\
      \cline{2-4} 

     &  & mse & 3784188.541 \\
     & \multirow{-2}{*}{baseline +  MLP} & mae & 1210.785 \\
     \cline{2-4} 

     &  & mse & \textbf{3421546.027} \\
    \multirow{-8}{*}{representation retrieval} & \multirow{-2}{*}{baseline + encoder} & mae & \textbf{1101.172} \\

     \midrule 
     
     &  & mse & 3421546.027 \\
     & \multirow{-2}{*}{RATSF with embedding} & mae & 1101.172 \\
     \cline{2-4} 

     &  & mse & \textbf{3343793.904} \\
    \multirow{-4}{*}{decoding} & \multirow{-2}{*}{RATSF with encoder} & mae & \textbf{1004.326} \\

    \bottomrule 
    \end{tabular}
}
\label{table_emb}
\end{table}

\subsubsection{Retrieval Embedding}
\label{section:exp_retrieve_emb}
\noindent
As previously mentioned, in our main experiments, we aimed to isolate the contribution of the encoder's output to the retrieval performance. To do this, we utilized the encoder's output $\mathbf{X}^e_r$ for the retrieval process and the embedding output $\mathbf{X}^b_r$ for decoding. In this section, we have completed the discussion of this part of the logic. Here, we will compare different schemes for retrieval and evaluate the effects of various decoding schemes.

We argue that DTW is not the optimal retrieval method and demonstrate this through an experiment on the FHSV dataset, where we compare DTW, a separately trained two-layer MLP, and our current design choice: using the forecasting model's encoder for retrieval. The upper half of Table \ref{table_emb} presents a comparison of the forecasting precision achieved with these different retrieval embedding strategies. The 'Baseline' indicates the standard Transformer model without any retrieval mechanism. As the results indicate, the encoder output outperforms the MLP, likely due to the attention mechanism within the Transformer's encoder, which is adept at identifying and refining the relationships among data points in sequences, thus capturing more accurate patterns.

The lower half of Table \ref{table_emb} compares the decoding performance of RACA when using the Transformer's embedding output and encoder output as the information source. The experimental results indicate that refining the retrieved sequence information with the Encoder can reduce the MAE by 18.38\% to 1004.326, compared to the vanilla Transformer, and by 8.79\% compared to using the embedding output. This substantiates the effectiveness of the RATSF architecture demonstrated in Figure \ref{fig1}.

\begin{table}[h] 
\centering
\caption{Effectiveness of Using DTW as Auxiliary Training.}
    \begin{tabular}{cccc}
    \toprule         
    dataset & number of epochs & \multicolumn{2}{c}{metric} \\
    \midrule 
     &  & mse & \textbf{3421546.027} \\
     & \multirow{-2}{*}{1 epoch} & mae & \textbf{1101.172} \\
     \cline{2-4} 
     &  & mse & 3769362.251 \\
     & \multirow{-2}{*}{2 epoch} & mae & 1252.136 \\
     \cline{2-4} 
     &  & mse & 3697616.753 \\
     & \multirow{-2}{*}{3 epoch} & mae & 1135.631 \\
     \cline{2-4} 
     &  & mse & 3720408.244 \\
     & \multirow{-2}{*}{4 epoch} & mae & 1248.433 \\
     \cline{2-4} 
     &  & mse & 3842794.029 \\
     & \multirow{-2}{*}{5 epoch} & mae & 1278.324 \\
     \cline{2-4} 
     &  & mse & 3800124.170 \\
    \multirow{-12}{*}{FHSV} & \multirow{-2}{*}{end to end} & mae & 1302.054 \\
    \bottomrule 
    \end{tabular}
\label{table7}
\end{table}

\subsubsection{Effectiveness of Using DTW as Auxiliary Training}
\label{section:dtw}
\noindent
In Section \ref{section:embedding learning}, we assert the effectiveness of employing DTW as a retrieval method during the cold start phase to facilitate the training of forecasting models for a single epoch. To substantiate the rationale for this strategy, we conducted a comparative experimental study: one baseline group without DTW-assisted training (referred to as 'end-to-end'), and several other groups, each trained with DTW assistance for 1 to 5 epochs. We ensured that all experiments had the same maximum number of training epochs and utilized an identical early stopping strategy. Table \ref{table7} demonstrates that the model's predictive performance is the weakest when DTW is not utilized, as indicated by the highest MSE and MAE values in the last row of Table \ref{table7}; furthermore, extending DTW-assisted training beyond one epoch did not lead to improved outcomes.

This result aligns with our hypothesis that in the absence of DTW support during the initial training phase, the model has difficulty identifying high-quality retrieval vectors, which are crucial for accurate forecasting. However, as our primary experiments have shown, DTW is not inherently the best retrieval method for enhancing predictive accuracy. Therefore, extending the period of DTW-assisted training may hinder rather than enhance the model's ability to express and forecast effectively.

\section{Conclusion}
\noindent
In conclusion, RATSF stands out in the domain of univariant time-series forecasting by integrating a TSKB that meticulously slices and indexes historical data, leading to superior retrieval precision.
This is further enhanced by the Encoder’s representations,
which act as a sophisticated retrieval mechanism. Coupled
with our innovative RACA mechanism for adeptly merging retrieved segments,
we achieve heightened forecast accuracy. The adaptability of
our method is highlighted by its compatibility with diverse
Transformer variants and its extensive, successful application within Fliggy’s service volume forecasting scenarios.

\bibliographystyle{ACM-Reference-Format}
\bibliography{sample}

\clearpage
\begin{appendices}
\section{Fliggy Hotel Service Volume Dataset}
\noindent
The Fliggy Hotel Service Volume Dataset (FHSV) is derived from service records within the Fliggy APP's customer service center, capturing the volume of daily customer service operations related to inquiries about hotel reservations, cancellations, and changes. Each instance of a service request received either through the APP interface or by phone is counted as a single service event. Daily aggregates of such events are recorded, forming individual data points. This dataset comprises 1740 data points collected between January 1, 2019, and October 7, 2023.

To facilitate experimentation, this time series dataset has been sequentially partitioned in a chronological order into three subsets: the train set (from January 1, 2019, up until May 24, 2022), the evaluation set (from May 25, 2022, to October 31, 2022), and the test set (from November 1, 2022, onward). The reported experimental results are based on the model's performance on the test set alone.

\begin{table}[h] 
\centering
\caption{Length of Retrieval Index Segment of Knowledge Base.}
    \begin{tabular}{cccc}
    \toprule         
    dataset & retrieval length & \multicolumn{2}{c}{metric} \\
    
    \midrule 
     &  & mse & 4621934.067 \\
     & \multirow{-2}{*}{4} & mae & 1303.329 \\
     \cline{2-4} 
     &  & mse & 3828479.700 \\
     & \multirow{-2}{*}{7} & mae & 1178.574 \\
     \cline{2-4} 
     &  & mse & { \textbf{3421546.027}} \\
     & \multirow{-2}{*}{14} & mae & { \textbf{1101.172}} \\
     \cline{2-4} 
     &  & mse & 3747974.421 \\
     & \multirow{-2}{*}{21} & mae & 1208.379 \\
     \cline{2-4} 
     &  & mse & 3816283.253 \\
     & \multirow{-2}{*}{28} & mae & 1243.133 \\
    \cline{2-4} 
     
     &  & mse & 3913793.800 \\
    \multirow{-12}{*}{FHSV} & \multirow{-2}{*}{35} & mae & 1332.472 \\
     
    \bottomrule 
    \end{tabular}
\label{table5}
\end{table}

\section{Other Experiment Details}
\subsection{Length of Retrieval Index Segment of Knowledge Base}

\noindent
The length of the retrieval segment in a knowledge base significantly affects the accuracy of retrieving the original sequence. Through controlled experiments, we identified the most suitable retrieval segment length for the Fliggy scenario. As shown in Table \ref{table5}, we incrementally extended the length of the retrieval segment \textbf{K} from 4 units to 35 (corresponding to half to five times the length of the prediction sequence at 7 units), then observed the impact on model performance as measured by MSE and MAE.

The results revealed that during the expansion of the retrieval segment from 4 to 14 units, the predictive performance improved. However, beyond this point, as the retrieval segment continued to lengthen, the predictive performance began to decline. This trend demonstrates that, in practical scenarios, as the retrieval segment grows from short to long, its representation transitions from being information-poor to increasingly mixed and complex.

This situation underscores why, in traditional time series databases, using full-ordered representations for retrieval often yields effective outcomes. By contrast, our \textbf{K},\textbf{V} design effectively circumvents this issue, providing flexible and accurate retrieval results.

\subsection{Optimal Retrieval Count}
\label{appendix:retrieve_cnt}
\noindent
Having established that integrating effective historical pieces positively impacts forecasting, we then inquire about the optimal number of sequences to retrieve. To demonstrate the effect of this choice, we gradually increase the number of integrated retrieve sequences from 0 to 5. As shown in Table \ref{table4}, during the process of expanding from integrating no historical piece to integrating up to three historical pieces, the MAE error consistently decreases; however, when the quantity of integrated historical pieces is further increased, the forecasting error starts to escalate.

Upon analyzing samples, the insight is obvious: not all of the top five retrieved historical piece for most of samples closely resemble the Ground Truth. The reason being, focusing solely on the TopK ranking without adequately considering similarity thresholds can easily lead to a scenario where, as K increases, the quality of the retrieved pieces becomes less assured and more prone to introduce noise and irrelevant information, thus potentially causing confusion instead.

\begin{table}[h] 
\centering
\caption{Results of recalling different numbers of pieces in our RATSF model.}
\begin{tabular}{cccc}
    \toprule         
    {dataset} & {methods} & \multicolumn{2}{c}{metric} \\
    \midrule 
    
     &  & mse & 4344450.512 \\
    \multirow{12}{*}{FHSV} & \multirow{-2}{*}{baseline} & mae & 1349.146 \\
    \cline{2-4} 

     &  & mse & 3678230.020 \\
     & \multirow{-2}{*}{recalled Top-1 pieces} & mae & 1160.101 \\
    \cline{2-4}

     &  & mse & 3700934.754 \\
     & \multirow{-2}{*}{recalled Top-2 pieces} & mae & 1155.952 \\
    \cline{2-4}

     &  & mse & {\bf 3421546.027} \\
     & \multirow{-2}{*}{recalled Top-3 pieces} & mae & {\bf 1101.172} \\
    \cline{2-4}

     &  & mse & 3460814.538 \\
     & \multirow{-2}{*}{recalled Top-4 pieces} & mae & 1108.895 \\
    \cline{2-4}

     &  & mse & 3750426.751 \\
    & \multirow{-2}{*}{recalled Top-5 pieces} & mae & 1231.564 \\ 
    \bottomrule 
\end{tabular}
\label{table4}
\end{table}

\end{appendices}

\end{document}